\PassOptionsToPackage{unicode}{hyperref}
\PassOptionsToPackage{hyphens}{url}
\PassOptionsToPackage{dvipsnames,svgnames,x11names}{xcolor}
\documentclass[
]{article}
\usepackage{amsmath,amssymb}
\usepackage{lmodern}
\usepackage{iftex}
\ifPDFTeX
  \usepackage[T1]{fontenc}
  \usepackage[utf8]{inputenc}
  \usepackage{textcomp} 
\else 
  \usepackage{unicode-math}
  \defaultfontfeatures{Scale=MatchLowercase}
  \defaultfontfeatures[\rmfamily]{Ligatures=TeX,Scale=1}
\fi
\IfFileExists{upquote.sty}{\usepackage{upquote}}{}
\IfFileExists{microtype.sty}{
  \usepackage[]{microtype}
  \UseMicrotypeSet[protrusion]{basicmath} 
}{}
\makeatletter
\@ifundefined{KOMAClassName}{
  \IfFileExists{parskip.sty}{%
    \usepackage{parskip}
  }{
    \setlength{\parindent}{0pt}
    \setlength{\parskip}{6pt plus 2pt minus 1pt}}
}{
  \KOMAoptions{parskip=half}}
\makeatother
\usepackage{xcolor}
\usepackage{longtable,booktabs,array}
\usepackage{calc} 
\usepackage{etoolbox}
\makeatletter
\patchcmd\longtable{\par}{\if@noskipsec\mbox{}\fi\par}{}{}
\makeatother
\IfFileExists{footnotehyper.sty}{\usepackage{footnotehyper}}{\usepackage{footnote}}
\makesavenoteenv{longtable}
\usepackage{graphicx}
\makeatletter
\def\maxwidth{\ifdim\Gin@nat@width>\linewidth\linewidth\else\Gin@nat@width\fi}
\def\maxheight{\ifdim\Gin@nat@height>\textheight\textheight\else\Gin@nat@height\fi}
\makeatother
\setkeys{Gin}{width=\maxwidth,height=\maxheight,keepaspectratio}
\makeatletter
\def\fps@figure{htbp}
\makeatother
\providecommand{\tightlist}{%
  \setlength{\itemsep}{0pt}\setlength{\parskip}{0pt}}
\NewDocumentCommand\citeproctext{}{}
\NewDocumentCommand\citeproc{mm}{%
  \begingroup\def\citeproctext{#2}\cite{#1}\endgroup}
\makeatletter
 \let\@cite@ofmt\@firstofone
 \def\@biblabel#1{}
 \def\@cite#1#2{{#1\if@tempswa , #2\fi}}
\makeatother
\newlength{\cslhangindent}
\newlength{\csllabelwidth}
\newenvironment{CSLReferences}[2] 
 {\begin{list}{}{%
  \setlength{\itemindent}{0pt}
  \setlength{\leftmargin}{0pt}
  \setlength{\parsep}{0pt}
  \ifodd #1
   \setlength{\leftmargin}{\cslhangindent}
   \setlength{\itemindent}{-1\cslhangindent}
  \fi
  \setlength{\itemsep}{#2\baselineskip}}}
 {\end{list}}
\usepackage{calc}

\ifLuaTeX
\usepackage[bidi=basic]{babel}
\else
\usepackage[bidi=default]{babel}
\fi
\babelprovide[main,import]{american}

\def\languageshorthands#1{}
\usepackage{pifont}

\ifLuaTeX
  \usepackage{selnolig}  
\fi
\IfFileExists{bookmark.sty}{\usepackage{bookmark}}{\usepackage{hyperref}}
\IfFileExists{xurl.sty}{\usepackage{xurl}}{} 
\hypersetup{
  pdftitle={LBR-Stack: ROS 2 and Python Integration of KUKA FRI for Med
and IIWA Robots},
  pdfauthor={Martin Huber, Christopher E. Mower, Sebastien Ourselin, Tom
Vercauteren, Christos Bergeles},
  pdflang={en-US},
  colorlinks=true,
  linkcolor={Maroon},
  filecolor={Maroon},
  citecolor={Blue},
  urlcolor={Blue},
  pdfcreator={LaTeX via pandoc}}

\title{LBR-Stack: ROS 2 and Python Integration of KUKA FRI for Med and
IIWA Robots}

\definecolor{c53baa1}{RGB}{83,186,161}
\definecolor{c202826}{RGB}{32,40,38}

\usepackage[affil-it]{authblk}
\usepackage{orcidlink}
\author[1%
  \ensuremath\mathparagraph]{Martin Huber%
    \,\orcidlink{0000-0003-4603-6773}\,%
    }
\author[1%
  ]{Christopher E. Mower%
    \,\orcidlink{0000-0002-3929-9391}\,%
    }
\author[1%
  ]{Sebastien Ourselin%
    \,\orcidlink{0000-0002-5694-5340}\,%
    }
\author[1%
  *%
  ]{Tom Vercauteren%
    \,\orcidlink{0000-0003-1794-0456}\,%
    }
\author[1%
  *%
  ]{Christos Bergeles%
    \,\orcidlink{0000-0002-9152-3194}\,%
    }

\affil[1]{School of Biomedical Engineering \& Imaging Sciences, King's
College London, United Kingdom%
  }
\affil[$\mathparagraph$]{Corresponding author: %
}
\affil[*]{These authors contributed equally.}
\date{21 November 2023}

\begin{document}
\maketitle

\begin{figure}
\centering
\includegraphics{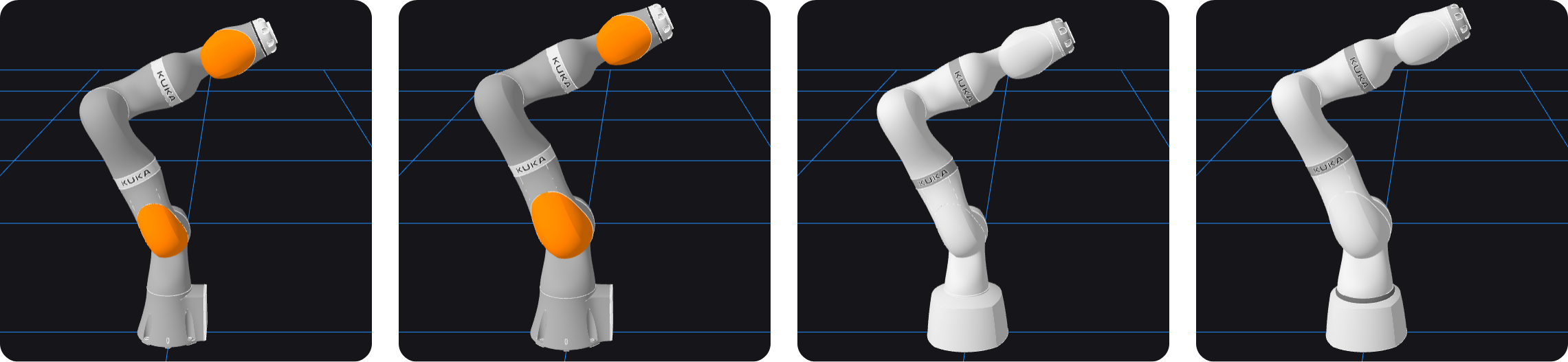}
\caption[Supported robots in the LBR-Stack. From left to right: KUKA LBR
IIWA 7 R800, IIWA 14 R820, Med 7 R800, Med 14 R820. Visualizations made
using Foxglove .]{Supported robots in the LBR-Stack. From left to right:
KUKA LBR IIWA 7 R800, IIWA 14 R820, Med 7 R800, Med 14 R820.
Visualizations made using Foxglove \footnotemark{}.}
\end{figure}
\footnotetext{Foxglove: \url{https://foxglove.dev/ros}.}

\section{Summary}\label{summary}

The \texttt{LBR-Stack} is a collection of packages that simplify the
usage and extend the capabilities of KUKA's Fast Robot Interface (FRI)
(\citeproc{ref-fri}{Schreiber et al., 2010}). It is designed for mission
critical hard real-time applications. Supported are the
\texttt{KUKA\ LBR\ Med\ 7/14} and \texttt{KUKA\ LBR\ IIWA\ 7/14} robots
in the Gazebo simulation (\citeproc{ref-gazebo}{Koenig \& Howard, 2004})
and for communication with real hardware. A demo video can be found
\href{https://drive.google.com/file/d/1_n3dFdFN74yWlijDQiGcNYg65NSG_ACs/view?usp=sharing}{here}.
An overview of the software architecture is shown in Figure
\ref{fig:fri}.

At the \texttt{LBR-Stack}'s core is the following package:

\begin{itemize}
\tightlist
\item
  \textbf{fri}: Integration of KUKA's original FRI client library into
  CMake: \href{https://github.com/lbr-stack/fri}{link}.
\end{itemize}

All other packages are built on top. These include Python bindings and
packages for integration into the Robot Operating System (ROS) and ROS
2:

\begin{itemize}
\tightlist
\item
  \textbf{pyfri}: Python bindings for the \textbf{fri}:
  \href{https://github.com/lbr-stack/pyfri}{link}.
\item
  \textbf{lbr\_fri\_ros2\_stack}: ROS 1/2 integration of the
  \texttt{KUKA\ LBR}s through the \textbf{fri}:
  \href{https://github.com/lbr-stack/lbr_fri_ros2_stack}{link}.
\end{itemize}

For brevity, and due to the architectural advantages over ROS
(\citeproc{ref-ros2}{Macenski et al., 2022}), only ROS 2 is considered
in the following. The \textbf{lbr\_fri\_ros2\_stack} comprises the
following packages:

\begin{itemize}
\tightlist
\item
  \textbf{lbr\_bringup}: Python library for launching the different
  components.
\item
  \textbf{lbr\_description}: Description files for the \texttt{Med7/14}
  and \texttt{IIWA7/14} robots.
\item
  \textbf{lbr\_demos}: Demonstrations for simulation and the real
  robots.
\item
  \textbf{lbr\_fri\_idl}: Interface Definition Language (IDL) equivalent
  of FRI protocol buffers.
\item
  \textbf{lbr\_fri\_ros2}: FRI ROS 2 interface through
  \texttt{realtime\_tools} (\citeproc{ref-ros_control}{Chitta et al.,
  2017}).
\item
  \textbf{lbr\_ros2\_control}: Interface and controllers for
  \texttt{ros2\_control} (\citeproc{ref-ros2_control}{Magyar et al.,
  2023}).
\item
  \textbf{lbr\_moveit\_config}: MoveIt 2 configurations
  (\citeproc{ref-moveit}{Coleman et al., 2014}).
\end{itemize}

\begin{figure}
\centering
\includegraphics{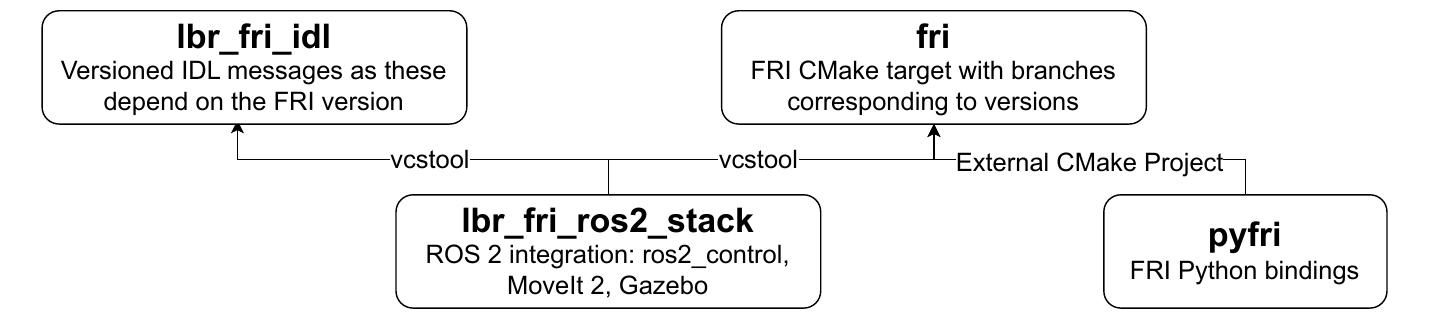}
\caption[An overview of the overall software architecture. There exists
a single source for KUKA's FRI. This design facilitates that downstream
packages, i.e.~the Python bindings and the ROS 2 package, can easily
support multiple FRI versions. The ROS 2 side utilizes
vcstool.\label{fig:fri}]{An overview of the overall software
architecture. There exists a single source for KUKA's FRI. This design
facilitates that downstream packages, i.e.~the Python bindings and the
ROS 2 package, can easily support multiple FRI versions. The ROS 2 side
utilizes vcstool\footnotemark{}.\label{fig:fri}}
\end{figure}
\footnotetext{vcstool: \url{https://github.com/dirk-thomas/vcstool}.}

\section{Statement of need}\label{statement-of-need}

An overview of existing work that interfaces the KUKA LBRs from an
external computer is given in Table 1. We broadly classify these works
into custom communication solutions
(\citeproc{ref-iiwa_stack}{Hennersperger et al., 2017};
\citeproc{ref-kuka_sunrise_toolbox}{Safeea \& Neto, 2019};
\citeproc{ref-libiiwa}{Serrano-Muñoz et al., 2023}) and communication
solutions through KUKA's FRI UDP channel
(\citeproc{ref-iiwa_ros2}{Bednarczyk \& Guzmán, 2023};
\citeproc{ref-iiwa_ros}{Chatzilygeroudis et al., 2019}). The former can
offer greater flexibility while the latter offer a well defined
interface and direct software support from KUKA. Contrary to the custom
communication solutions, the FRI solutions additionally enable hard
real-time communication, that is beneficial for mission critical
development. Stemming from translational medical research, this work
therefore focuses on the FRI.

Limitations with the current FRI solutions are:

\begin{enumerate}
\def\labelenumi{\arabic{enumi}.}
\tightlist
\item
  Only support \texttt{IIWA\ 7/14} robots, not \texttt{Med\ 7/14}.
\item
  Don't provide Python bindings.
\item
  Don't support multiple FRI versions:

  \begin{itemize}
  \tightlist
  \item
    Modified FRI client source code
    \href{https://github.com/epfl-lasa/iiwa_ros}{iiwa\_ros}.
  \item
    FRI client library tangled into the source code
    \href{https://github.com/ICube-Robotics/iiwa_ros2}{iiwa\_ros2}.
  \end{itemize}
\item
  Partial support of FRI functionality. Both,
  \href{https://github.com/epfl-lasa/iiwa_ros}{iiwa\_ros} and
  \href{https://github.com/ICube-Robotics/iiwa_ros2}{iiwa\_ros2},
  exclusively aim at providing implementations of the ROS 1/2 hardware
  abstraction layer. This does not support:

  \begin{itemize}
  \tightlist
  \item
    FRI's cartesian impedance control mode.
  \item
    FRI's cartesian control mode (FRI version 2 and above).
  \end{itemize}
\end{enumerate}

The first original contribution of this work is to add support for the
\texttt{KUKA\ LBR\ Med\ 7/14} robots, which, to the best author's
knowledge, does not exist in any other work. The second novel
contribution of this work is to provide Python bindings. This work
solves the support for multiple FRI versions by treating the FRI library
as an externally provided library by separating it into the \textbf{fri}
package, which leaves the FRI's source code untouched and simply
provides build support. The partial support for the FRI functionality is
solved by defining an IDL message to KUKA's \texttt{nanopb} command and
state protocol buffers in \textbf{lbr\_fri\_idl}. These messages can
then be interfaced from ROS 1/2 topics via simple controllers or from
the ROS 1/2 hardware abstraction layer.

\begin{longtable}[]{@{}
  >{\raggedright\arraybackslash}p{(\columnwidth - 22\tabcolsep) * \real{0.2419}}
  >{\raggedright\arraybackslash}p{(\columnwidth - 22\tabcolsep) * \real{0.0645}}
  >{\raggedright\arraybackslash}p{(\columnwidth - 22\tabcolsep) * \real{0.0645}}
  >{\raggedright\arraybackslash}p{(\columnwidth - 22\tabcolsep) * \real{0.0484}}
  >{\raggedright\arraybackslash}p{(\columnwidth - 22\tabcolsep) * \real{0.0968}}
  >{\raggedright\arraybackslash}p{(\columnwidth - 22\tabcolsep) * \real{0.0323}}
  >{\raggedright\arraybackslash}p{(\columnwidth - 22\tabcolsep) * \real{0.0484}}
  >{\raggedright\arraybackslash}p{(\columnwidth - 22\tabcolsep) * \real{0.0806}}
  >{\raggedright\arraybackslash}p{(\columnwidth - 22\tabcolsep) * \real{0.0484}}
  >{\raggedright\arraybackslash}p{(\columnwidth - 22\tabcolsep) * \real{0.0484}}
  >{\raggedright\arraybackslash}p{(\columnwidth - 22\tabcolsep) * \real{0.1290}}
  >{\raggedright\arraybackslash}p{(\columnwidth - 22\tabcolsep) * \real{0.0968}}@{}}
\caption{Overview of existing frameworks for interfacing the KUKA LBRs.
A square indicates support for the respective feature. List of
abbreviations: Hard Real-time (\textbf{RT}), Position Control
(\textbf{Pos}), Impedance Control (\textbf{Imp}), Cartesian Impedance
Control (\textbf{Cart Imp}), Hardware Interface (\textbf{HW
IF}).}\tabularnewline
\toprule\noalign{}
\begin{minipage}[b]{\linewidth}\raggedright
Framework
\end{minipage} & \begin{minipage}[b]{\linewidth}\raggedright
IIWA
\end{minipage} & \begin{minipage}[b]{\linewidth}\raggedright
Med
\end{minipage} & \begin{minipage}[b]{\linewidth}\raggedright
ROS
\end{minipage} & \begin{minipage}[b]{\linewidth}\raggedright
ROS 2
\end{minipage} & \begin{minipage}[b]{\linewidth}\raggedright
RT
\end{minipage} & \begin{minipage}[b]{\linewidth}\raggedright
FRI
\end{minipage} & \begin{minipage}[b]{\linewidth}\raggedright
pyfri
\end{minipage} & \begin{minipage}[b]{\linewidth}\raggedright
Pos
\end{minipage} & \begin{minipage}[b]{\linewidth}\raggedright
Imp
\end{minipage} & \begin{minipage}[b]{\linewidth}\raggedright
Cart Imp
\end{minipage} & \begin{minipage}[b]{\linewidth}\raggedright
HW IF
\end{minipage} \\
\midrule\noalign{}
\endfirsthead
\toprule\noalign{}
\begin{minipage}[b]{\linewidth}\raggedright
Framework
\end{minipage} & \begin{minipage}[b]{\linewidth}\raggedright
IIWA
\end{minipage} & \begin{minipage}[b]{\linewidth}\raggedright
Med
\end{minipage} & \begin{minipage}[b]{\linewidth}\raggedright
ROS
\end{minipage} & \begin{minipage}[b]{\linewidth}\raggedright
ROS 2
\end{minipage} & \begin{minipage}[b]{\linewidth}\raggedright
RT
\end{minipage} & \begin{minipage}[b]{\linewidth}\raggedright
FRI
\end{minipage} & \begin{minipage}[b]{\linewidth}\raggedright
pyfri
\end{minipage} & \begin{minipage}[b]{\linewidth}\raggedright
Pos
\end{minipage} & \begin{minipage}[b]{\linewidth}\raggedright
Imp
\end{minipage} & \begin{minipage}[b]{\linewidth}\raggedright
Cart Imp
\end{minipage} & \begin{minipage}[b]{\linewidth}\raggedright
HW IF
\end{minipage} \\
\midrule\noalign{}
\endhead
\bottomrule\noalign{}
\endlastfoot
\href{https://github.com/lbr-stack}{lbr-stack} & \(\bullet\) &
\(\bullet\) & \(\bullet\) & \(\bullet\) & \(\bullet\) & \(\bullet\) &
\(\bullet\) & \(\bullet\) & \(\bullet\) & \(\bullet\) & \(\bullet\) \\
\href{https://github.com/epfl-lasa/iiwa_ros}{iiwa\_ros} & \(\bullet\) &
& \(\bullet\) & & \(\bullet\) & \(\bullet\) & & \(\bullet\) &
\(\bullet\) & & \(\bullet\) \\
\href{https://github.com/ICube-Robotics/iiwa_ros2}{iiwa\_ros2} &
\(\bullet\) & & & \(\bullet\) & \(\bullet\) & \(\bullet\) & &
\(\bullet\) & \(\bullet\) & & \(\bullet\) \\
\href{https://github.com/IFL-CAMP/iiwa_stack}{iiwa-stack} & \(\bullet\)
& & \(\bullet\) & & & & & \(\bullet\) & \(\bullet\) & \(\bullet\) & \\
\href{https://github.com/Toni-SM/libiiwa}{libiiwa} & \(\bullet\) & &
\(\bullet\) & \(\bullet\) & & & & \(\bullet\) & \(\bullet\) &
\(\bullet\) & \\
\href{https://github.com/Modi1987/KST-Kuka-Sunrise-Toolbox}{KST-KUKA} &
\(\bullet\) & & & & & & & \(\bullet\) & \(\bullet\) & \(\bullet\) & \\
\end{longtable}

\section{Acknowledgement}\label{acknowledgement}

We want to acknowledge the work in
(\citeproc{ref-iiwa_stack}{Hennersperger et al., 2017}), as their MoveIt
configurations were utilized in a first iteration of this project.

This work was supported by core funding from the Wellcome/EPSRC
{[}WT203148/Z/16/Z; NS/A000049/1{]}, the European Union's Horizon 2020
research and innovation programme under grant agreement No 101016985
(FAROS project), and EPSRC under the UK Government Guarantee Extension
(EP/Y024281/1, VITRRO).

\section{References}\label{references}

\phantomsection\label{refs}


\begin{CSLReferences}{1}{0}
\bibitem[\citeproctext]{ref-iiwa_ros2}
Bednarczyk, M., \& Guzmán, J. H. G. (2023). ROS 2 stack for KUKA iiwa
collaborative robots. In \emph{GitHub repository}. GitHub.
\url{https://github.com/ICube-Robotics/iiwa_ros2}

\bibitem[\citeproctext]{ref-iiwa_ros}
Chatzilygeroudis, K., Mayr, M., Fichera, B., \& Billard, A. (2019).
\emph{Iiwa\_ros: A ROS stack for KUKA's IIWA robots using the fast
research interface}. \url{http://github.com/epfl-lasa/iiwa_ros}

\bibitem[\citeproctext]{ref-ros_control}
Chitta, S., Marder-Eppstein, E., Meeussen, W., Pradeep, V., Rodríguez
Tsouroukdissian, A., Bohren, J., Coleman, D., Magyar, B., Raiola, G.,
Lüdtke, M., \& Fernández Perdomo, E. (2017). Ros\_control: A generic and
simple control framework for ROS. \emph{The Journal of Open Source
Software}. \url{https://doi.org/10.21105/joss.00456}

\bibitem[\citeproctext]{ref-moveit}
Coleman, D., Sucan, I., Chitta, S., \& Correll, N. (2014). Reducing the
barrier to entry of complex robotic software: A moveit! Case study.
\emph{arXiv Preprint arXiv:1404.3785}.
\url{https://doi.org/10.6092/JOSER_2014_05_01_p3}

\bibitem[\citeproctext]{ref-iiwa_stack}
Hennersperger, C., Fuerst, B., Virga, S., Zettinig, O., Frisch, B.,
Neff, T., \& Navab, N. (2017). Towards MRI-based autonomous robotic US
acquisitions: A first feasibility study. \emph{IEEE Transactions on
Medical Imaging}, \emph{36}(2), 538--548.
\url{https://doi.org/10.1109/TMI.2016.2620723}

\bibitem[\citeproctext]{ref-gazebo}
Koenig, N., \& Howard, A. (2004). Design and use paradigms for gazebo,
an open-source multi-robot simulator. \emph{2004 IEEE/RSJ International
Conference on Intelligent Robots and Systems (IROS) (IEEE Cat.
No.04CH37566)}, \emph{3}, 2149--2154 vol.3.
\url{https://doi.org/10.1109/IROS.2004.1389727}

\bibitem[\citeproctext]{ref-ros2}
Macenski, S., Foote, T., Gerkey, B., Lalancette, C., \& Woodall, W.
(2022). Robot operating system 2: Design, architecture, and uses in the
wild. \emph{Science Robotics}, \emph{7}(66).
\url{https://doi.org/10.1126/scirobotics.abm6074}

\bibitem[\citeproctext]{ref-ros2_control}
Magyar, B., Stogl, D., Knese, K., \& Community. (2023). Generic and
simple controls framework for ROS 2. In \emph{GitHub repository}.
GitHub. \url{https://github.com/ros-controls/ros2_control}

\bibitem[\citeproctext]{ref-kuka_sunrise_toolbox}
Safeea, M., \& Neto, P. (2019). KUKA sunrise toolbox: Interfacing
collaborative robots with MATLAB. \emph{IEEE Robotics Automation
Magazine}, \emph{26}(1), 91--96.
\url{https://doi.org/10.1109/MRA.2018.2877776}

\bibitem[\citeproctext]{ref-fri}
Schreiber, G., Stemmer, A., \& Bischoff, R. (2010). The fast research
interface for the kuka lightweight robot. \emph{IEEE Workshop on
Innovative Robot Control Architectures for Demanding (Research)
Applications How to Modify and Enhance Commercial Controllers (ICRA
2010)}, 15--21.

\bibitem[\citeproctext]{ref-libiiwa}
Serrano-Muñoz, A., Elguea-Aguinaco, Í., Chrysostomou, D., BØgh, S., \&
Arana-Arexolaleiba, N. (2023). A scalable and unified multi-control
framework for KUKA LBR iiwa collaborative robots. \emph{2023 IEEE/SICE
International Symposium on System Integration (SII)}, 1--5.
\url{https://doi.org/10.1109/SII55687.2023.10039308}

\end{CSLReferences}
\end{document}